# Query Chains: Learning to Rank from Implicit Feedback

Filip Radlinski
Department of Computer Science
Cornell University
Ithaca, NY, USA
filip@cs.cornell.edu

Thorsten Joachims
Department of Computer Science
Cornell University
Ithaca, NY, USA
tj@cs.cornell.edu

## ABSTRACT

This paper presents a novel approach for using clickthrough data to learn ranked retrieval functions for web search results. We observe that users searching the web often perform a sequence, or chain, of queries with a similar information need. Using query chains, we generate new types of preference judgments from search engine logs, thus taking advantage of user intelligence in reformulating queries. To validate our method we perform a controlled user study comparing generated preference judgments to explicit relevance judgments. We also implemented a real-world search engine to test our approach, using a modified ranking SVM to learn an improved ranking function from preference data. Our results demonstrate significant improvements in the ranking given by the search engine. The learned rankings outperform both a static ranking function, as well as one trained without considering query chains.

## Categories and Subject Descriptors

H.3.3 [**Information Storage and Retrieval**]: Information Search and Retrieval

## 1. INTRODUCTION

Designing effective ranking functions for free text retrieval has proved notoriously difficult. Retrieval functions designed for one collection and application often do not work well on other collections without additional time consuming modifications. This has led to interest in using machine learning methods for automatically learning ranked retrieval functions.

For this learning task, training data can be collected in two ways. One approach relies on actively soliciting training data by recording user queries and then asking users to explicitly provide relevance judgments on retrieved documents (such as [7, 13, 22]). Few users are willing to do this, making significant amounts of such data difficult to obtain. An alternative approach is to extract implicit relevance feedback from search engine log files (such as in [6, 15]). This allows virtually unlimited data to be collected at very low cost, although interpretation is more complex.

Irrespective of the approach, to the best of our knowledge all previous research in learning retrieval functions has considered each query independently. We will show that this ignores valuable information that is hidden in the sequence of queries and clicks in a search session. For instance, if we repeatedly observe the query "special collections" followed by another for "rare books" on a library search system, we may deduce that web pages relevant to the second query may also be relevant to the first. Additionally, this log information can also allow us to learn to correct spelling mistakes in a similar way. For example, we observed that users searching for the "Lexis Nexis" repository often first search for "Lexis Nexus" by mistake.

As users search, it is well documented that they often reformulate their queries [3, 8, 18, 20]. Previous work has attempted to predict query reformulations, but to the best of our knowledge these reformulations have never been used to learn better retrieval functions. In this paper, we refer to a sequence of reformulated queries as a *query chain*. When queries are considered independently, log files only provide implicit feedback on a few results at the top of the result set for each query because users very rarely look further down the list. The advantage of using query chains is that we can also deduce relevance judgments on the many more documents seen during an entire search session.

The key contribution of this work is recognizing that we can successfully use evidence of query chains that is present in search engine log files to learn better retrieval functions. We demonstrate a simple method for automatically detecting query chains in query and clickthrough logs. Using this data, we show how to infer preference judgments as to the relative relevance of documents both within individual query results, and between documents returned by different queries within the same query chain. The method used to generate the preference judgments is validated using a controlled user study. We then adapt a ranking SVM to learn a ranked retrieval function from the preference judgments. In doing so, we propose a general retrieval model that can learn to associate individual documents with specific query words, even if the words do not occur in the documents. This differs from previous learned ranked retrieval functions in that our method can learn a much more general class of functions.

We demonstrate the effectiveness of our approach on a real-world web search system, the Cornell University library[1] web search. We name our implementation the Osmot search engine, and it is available for download to the research community. The name is derived from the word *osmosis*, as learning from implicit feedback is, in our opinion, almost as good as learning from users by osmosis.

## 2. RELATED WORK

When learning to rank, the method by which training data is collected offers an important way to distinguish between different approaches. This data usually consists of a set of statements as to the relevance of a document, or set of documents, to a given query. Such relevance judgments are either collected explicitly by asking users, or implicitly by observing user behavior and drawing conclusions. Moreover, the statements can be absolute or relative. Absolute feedback involves statements that a particular document is, or is not, relevant to a query. Relative feedback involves statements that a particular document is more relevant to a query than some other document.

Most previous work in learning to rank has assumed absolute relevance judgments. On the one hand, a number of methods in ordinal regression use explicit feedback to learn to rank, such as work by Crammer and Singer [7], Rajaram et al. [22] and Herbrich et al. [13]. However, explicit feedback is expensive to collect, with few users willing to spend the additional time to provide it in a real-world setting. This makes typical labeled data sets small and difficult to work with. A number of researchers have collected absolute relevance judgments implicitly from clickthrough logs, such as [4, 6, 19, 25]. They postulate that documents clicked on in search results are highly likely to be relevant. For example, Kemp et al. [19] present a learning search engine using document transformation. They assume results clicked on are relevant to the query and append the query to these documents. However, implicit clickthrough data has been shown to be biased as it is relative to the retrieval function quality and ordering [15, 17]. This makes its interpretation as absolute feedback of questionable accuracy.

Cohen et al. [6] and Freund et al. [9] propose using log data to generate relative preference feedback. Both approaches consider learning a ranking function from these preference judgments, along similar lines as this work. However, in contrast to our method their learned function is limited to a combination of rankings given by a fixed set of manually constructed "experts". This approach of learning a combination of functions is also used by most other work in this area [1, 2, 4, 15, 21].

Joachims [15] refined the interpretation of clickthrough log data as relative feedback. He suggests that given a ranking and a clicked-on document $d$, any document ranked above $d$ but not clicked on is likely less relevant than $d$. In this paper, we evaluate the validity of this construction, and extend it to query chains. We also use a more general ranking function and extend the learning algorithm to query chains.

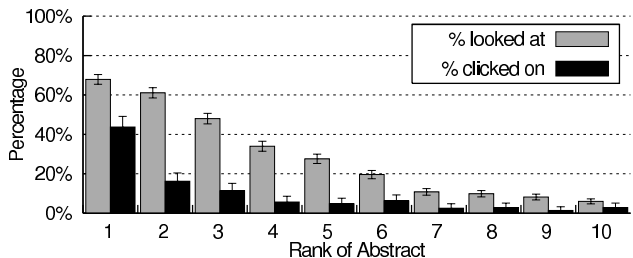

**Figure 1: Percentage of time an abstract was viewed/clicked on depending on the rank of the result.**

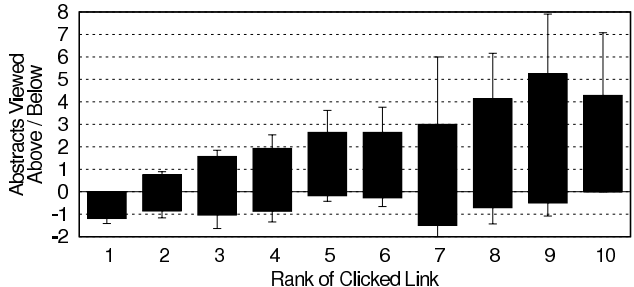

**Figure 2: Mean number of abstracts viewed above and below a clicked link depending on its rank.**

An important innovation in this paper is that we learn a more general ranking function than previous work by associating query words with specific documents. This approach has been used previously to learn to generate abstracts [23], and in document transformation [19], but not to learn ranking functions. Prior approaches cannot learn to associate "new" documents with a given query because they combine or re-order results obtained from one or more static ranking functions. In particular, given a query $q$, they cannot learn to retrieve any document not originally returned by $q$. Coming closest to solving this limitation previously, the method presented by Kemp et al. [19] could be extended with query chains. However, they assume implicit absolute feedback, making their approach more likely to be susceptible to bias and noise.

## 3. ANALYSIS OF USER BEHAVIOR

In order to infer implicit preference judgments from log files, we need to understand how users assess search results. Clearly we can only derive valid feedback for results that the user actually looked at and assessed. In this section we explore this question.

An eye tracking study was performed to observe how users formulate queries, assess the results returned by the search engine and select the links they click on [11, 12]. Thirty six undergraduate student volunteers were instructed to search for the answers to five navigational and five informational queries [5]. The former involved finding a specific web page while the latter involved finding some specific information. The subjects were asked to start from the Google search page and find the answers. There were no restrictions on what queries they may choose, how and when to reformulate queries, or which links to follow. Users were told that the goal of the study was to observe how people search the

| Query 1: NDLF |
| --- |
| 1. http://.../staffweb/SMG/SMG970319.html |
| 2. http://.../staffweb/SMG/SMG970226.html |
| 3. http://.../staffweb/SMG/SMG960417.html |
| 4. http://.../staffweb/SMG/SMG960403.html |
| 5. http://.../staffweb/SMG/SMG960828.html |

| Query 2: "Ezra Cornell" residence |
| --- |
| 1. Dear Uncle Ezra – Questions for Tuesday, May... |
| 2. Dear Uncle Ezra – Questions for Thursday,... |
| 3. Ezra Cornell had close Albion ties |
| 4. October 1904 – Albion 100 Years Age |
| 5. Cornell competes with Off-Housing market |
| ⋮ |

**Figure 3: Two example queries and result sets.**

Web, but were not told of the specific interest in their behavior on the results page of Google. All clicks, the results returned by Google, and the pages connected to the results were recorded by an HTTP proxy. Movement of the eyes was recorded using an ASL 504 commercial eye tracker (Applied Science Technologies, Bedford, MA). More details on the experimental setup are provided in [12].

Figure 1 shows the fraction of the time users looked at, and clicked on, each of the top 10 search results for a query. It tells us that users usually look at least at the top two result abstracts. Interestingly, note that despite the top two documents receiving almost equal attention, users were much more likely to click on the first result. Figure 2 (adapted from Figure 2 in [12]) shows the number of abstracts viewed above and below any result that was clicked on. This figure tells us that users usually scan the results in order from top to bottom. We also see that users usually look at one abstract below any they click on. Further analysis showed that this is usually the abstract immediately below the one clicked on [17]. We conclude that users typically look at most of the results from the first to the one below the last one clicked on.

Previous work studying web search behavior [20, 24] observed that users rarely run only a single query and immediately find suitable results. Rather, they tend to perform a sequence of queries for any given question. Such query chains are also observed in the eye tracking study. The mean query chain length was 2.2 queries, although the particular questions asked and the laboratory environment would be expected to have an influence on this value. A number of papers (e.g. [3, 10, 18]) successfully learn to predict query reformulations. Their success on this task suggests that the problem of detecting query chains, which we will have to address, is feasible.

# 4. FROM LOG FILES TO FEEDBACK

This section details our approach for generating relative preference feedback from query and clickthrough logs as implemented in the Osmot search engine. We then present an evaluation of this approach using results from the eye tracking study.

Consider the queries shown in Figure 3 as examples we use to demonstrate the value of query chains. The first shows the results presented to a user running the query "NDLF" on the Cornell University library search page. The user is searching for the National Digital Library Foundation website, but has retrieved only meeting notes that reference people working for the NDLF. The desired page is not in these results, most probably because it does not contain the word "NDLF". The second query is a search performed in Google by a participant in the eye tracking study in attempting to find the name of the house that Ezra Cornell built for himself. We get many results, but in fact none of the top 10 contain any relevant information. In both cases, single query feedback will not be informative because no relevant documents were retrieved. In the former case, the results simply do not contain any documents relevant to the query. In the latter, if there is a relevant document it is unlikely the user will look far enough in the results to see it.

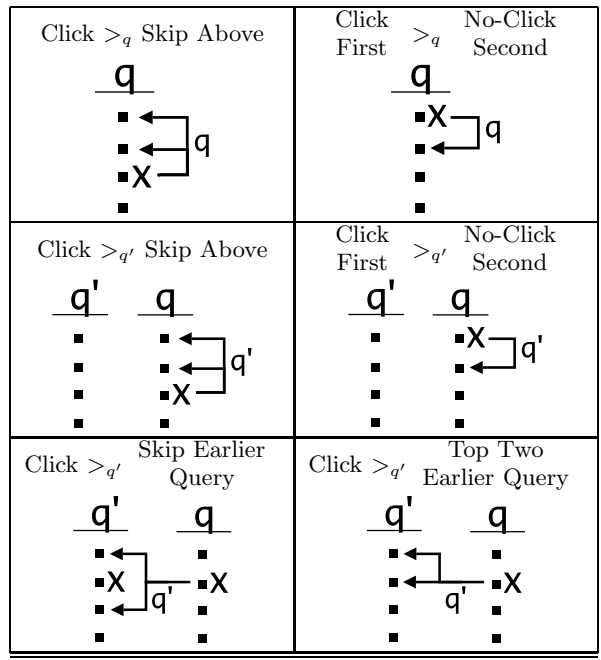

**Figure 4: Feedback strategies. We either consider a single query, $q$, or a query $q$ that has been preceded by a query $q'$. Given a query, a dot represents a result document and an $x$ indicates the result was clicked on. We generate a constraint for each arrow shown, with respect to the query marked.**

On the other hand, after both of these queries, we observed that the user continued running other queries. Often, such later queries are more successful. If a user finds a relevant document with a later query, it is reasonable to assume that the user would have preferred to have seen the relevant document over the results actually returned earlier. Recognizing the information necessary to make these deductions is present in search engine log files, we now propose specific strategies for generating such preference feedback from query chains. We defer a discussion of how to group queries into query chains to Section 6.

## 4.1 Implicit Feedback Strategies

We generate preference feedback using six strategies. These strategies are illustrated in Figure 4. The first two strategies show preferences that can be inferred without query chains.

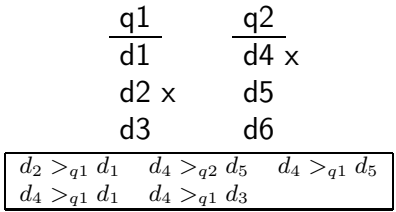

| q1 | q2 |
|----|----|
| d1 | d4 x |
| d2 x | d5 |
| d3 | d6 |

| $d_2 >_{q1} d_1$ | $d_4 >_{q2} d_5$ | $d_4 >_{q1} d_5$ |
|---|---|---|
| $d_4 >_{q1} d_1$ | $d_4 >_{q1} d_3$ | |

**Figure 5: Sample query chain and the feedback that would be generated using all six feedback strategies. Two queries were run, and each returned three documents. One document in each query was clicked on. $d_i >_q d_j$ means that $d_i$ is preferred over $d_j$ with respect to the query $q$.**

| Strategy | Accuracy |
|----------|----------|
| Click $>_q$ Skip Above | 78.2 ± 5.6 |
| Click First $>_q$ No-Click Second | 63.4 ± 16.5 |
| Click $>_q$ Skip Earlier Query | 68.0 ± 8.4 |
| Click $>_q$ Top Two Earlier Query | 84.5 ± 6.1 |
| Inter-Judge Agreement | 86.4 |

**Table 1: Accuracy of the strategies for generating pairwise preferences from clicks. The base of comparison are the explicit page judgments. Note that the first two cases cover two preferences strategies each.**

gives a sample query chain and the feedback that would be generated in this case.

## 4.2 Accuracy of Feedback Strategies

While the feedback strategies proposed above are intuitively appealing, a quantitative evaluation is necessary to establish their degree of validity. To determine the accuracy of each individual strategy, we conducted a controlled experiment following the setup of the eye-tracking study described in Section 3 for an additional 16 subjects. For these subjects, we evaluated in how far the preferences derived from the feedback strategies agree with explicit relevance judgments made by independent judges.

For these 16 subjects, we collected all results and their associated web pages returned by Google from the HTTP-proxy cache that recorded their sessions. We grouped the results by query chain and subject and collected explicit relevance judgments using five judges. The judges were asked to weakly order all results encountered during each query chain according to their relevance to the question. To avoid biasing the judges, the order in which results were presented to the judges was randomized and the judges were not given the abstracts Google used when presenting the results. Some of the query chains were assessed by two judges for inter-judge agreement verification. The agreement between judges is reasonably high. Whenever two judges expressed a strict preference between two pages, they agree in the direction of preference in 86.4% of the cases.

We now evaluate the extent to which the preferences generated from clicks agree with the explicit judgments. Table 1 summarizes the results. The table shows the percentage of times the preferences generated from clicks using the above strategies agree with the direction of a strict preference of a relevance judge. The first two lines in the table show the accuracy of the strategies that do not exploit query chains. The "Click $>_q$ Skip Above" strategy is 78.2% accurate, which is substantially and significantly better than the random baseline of 50%. Furthermore, it is reasonably close in accuracy to the average agreement of 86.4% between the explicit judgments from different judges, which can serve as an upper bound for the accuracy one could ideally expect even from explicit user feedback. The second within-query strategy, "Click First $>_q$ No-Click Second", appears less accurate. However, since it produces fewer preferences (i.e. only on queries where the user clicked exclusively on the first link), the confidence intervals are large. Independent of the accuracy, the preferences from this strategy are probably less informative, since they only confirm the current ranking and never suggest a reordering.

The first one, "Click $>_q$ Skip Above" was proposed in [6, 15]. This strategy proposes that given a clicked-on document (marked $x$ in the figure), any higher ranked document that was not clicked on is likely less relevant. The preference is indicated by an arrow labeled with the query, to show that the preference is with respect to that query. We expect this to be valid because the eye tracking study showed that users view results in order, and a user is unlikely to click on a document she considers less relevant than another document she observed. Note that these preferences are not stating that the clicked-on document *is* relevant, rather that it is *more likely* to be relevant than the ones not clicked on above. The second strategy, "Click First $>_q$ No-Click Second" makes use of the fact that users typically view both of the top two results before clicking. It states that if the first document is clicked on, but the second is not, the first is likely more relevant than the second. It seems reasonable to assume that having considered two options, the user is likely to click on the more relevant one.

The next two strategies are identical to the first two except that they generate feedback with respect to the *previous* query. The intuition behind this is that since the two queries belong to the same query chain, the user is looking for the same information with both. Had the user been presented with the new results for the earlier query, she would have preferred the clicked-on document over those skipped above.

The last two strategies make the most use of query chains. The strategy "Click $>_{q'}$ Skip Earlier Query" states that a clicked-on document is preferred over any result not clicked on in an earlier query $q'$ (within the same query chain). This judgment is made with respect to the earlier query,[2] $q'$. Since the eye tracking study revealed that users usually look one document past the last one clicked on, we also generate a preference for this document. In the event that no documents were clicked on in the earlier query, we use the fact that users usually look at the top two results. This is exploited in the feedback strategy "Click $>_{q'}$ Top Two Earlier Query" by generating preferences for the top two results. In the unusual case where there are not enough results to the earlier query to use these strategies, we select a random document as if it had been at the top two results.

Ultimately, given some query chain, we make use of all six strategies to generate the preference feedback. Figure 5

---

[2]It is unnecessary to state the same thing with respect to the later query $q$ because presumably the preference is already satisfied, or the user would have seen the same result earlier.

Lines 3 and 4 in Table 1 show the accuracy of the two strategies that exploit query chains. Both "Click $>_{q'}$ Skip Earlier Query" and "Click $>_{q'}$ Top Two Earlier Query" are significantly more accurate than random. In particular, the accuracy of "Click $>_{q'}$ Top Two Earlier Query" is very close to the average agreement between judges. Note that this strategy produces particularly informative preferences, since it associates documents with query words that may not occur in the document.

A possible explanation for the difference in accuracy between the two query-chain strategies is that they apply to different types of query chains. While "Click $>_{q'}$ Skip Earlier Query" is applied when the previous query received a click, the strategy "Click $>_{q'}$ Top Two Earlier Query" is applied precisely in the opposite case. To investigate the effect of this difference, we also evaluated a variant of "Click $>_{q'}$ Top Two Earlier Query". This variant generates preferences analogous to "Click $>_{q'}$ Top Two Earlier Query", but in chains where the previous query did receive a click (but excluding the clicked results). The accuracy of this strategy is $67.7\% \pm 9.4$, indicating that the absence of a click followed by another query with a click is particularly strong evidence regarding the relevance of the results of the earlier query.

Overall, we conclude that the preferences generated from the clickthrough logs are reasonably accurate and that they convey information regarding the user's preferences.

## 5. EVALUATION ENVIRONMENT

While the previous section showed that the preferences generated from logs files are accurate, can they be used to learn an improved retrieval system?

To address this question, we constructed a publicly accessible real-world search engine. The search engine implements a full-text search of web pages maintained by the Cornell University library[1] (CUL). This collection includes over 13,500 web pages. We used the Nutch search engine[3] as a starting point, with the Osmot search engine effectively being a wrapper around Nutch that implements logging, log analysis, learning, reranking and evaluation functionality. Osmot is designed to allow any number of different ranking functions to be plugged into it. In the experiments in this paper, we chose Nutch's built-in retrieval function as the baseline to compare against and build upon. The Nutch retrieval function is based on the cosine distance and incorporates several modifications to make it more suitable for web search including special cases for phrase matches and HTML fields.

## 6. DETECTING QUERY CHAINS

In order to use query chains, we must first have a method to identify them. In this section we propose such a heuristic and demonstrate its effectiveness.

As a basis for our evaluation, we created a dataset using search logs from the CUL search engine. We manually labeled query chains in the logs for a period of 5 weeks. The search logs recorded the query, date, IP address, results returned, number of clicks on the results and a session id uniquely assigned to each user. We extracted the list of queries, grouped them by IP address and sorted them chronologically. Queries from an IP address with no other queries within 24 hours were automatically marked as not

---

[3] http://www.nutch.org/

---

| CosineDistance(q1, q2) |
| CosineDistance(doc ids of r1', doc ids of r2') |
| CosineDistance(abstracts of r1', abstracts of r2') |
| TrigramMatch(q1, q2) |
| ShareOneWord(q1, q2) |
| ShareTwoWords(q1, q2) |
| SharePhraseOfTwoWords(q1, q2) |
| NumberOfDifferentWords(q1, q2) |
| $t2 - t1 \leq \{5, 10, 30, 100\}$ seconds |
| $t2 - t1 > 100$ seconds |
| NormalizedNumberOfClicks(r1) |
| NormalizedMin($|r1|$, $|r2|$) |
| NormalizedMax($|r1|$, $|r2|$) |

**Table 2: Features used to learn to classify query chains.** $q1$ and $q2$ are two queries at times $t1$ and $t2$, with $t1 < t2$. $r1$ and $r2$ are the respective result sets, with $r1'$ and $r2'$ being the top 10 results.

belonging to a query chain. This resulted in 1285 queries. Two judges (the authors of this paper) then individually grouped the queries into query chains manually, using search engines to resolve uncertainties (such as a query for a person followed by one for the department where the person is a faculty member). Finally, the judges combined their identified query chains, resolving the small number of disagreements between themselves through further investigation.

For each pair of queries from the same IP address within half an hour, we generated a training example by constructing a feature vector. The training example was labeled using the query chains identified manually. If the two queries belonged to the same query chain the example was labeled as positive. Otherwise it was labeled as negative. This led to 3418 training examples of which 3096 were labeled as positive. The feature vector generated given two queries $q1$ and $q2$ consisted of the 16 features shown in Table 2.

Using this data, we trained a number of SVM classifiers with various parameters. The classifiers learned tended to label almost all examples as positive. Among our best performing models was an SVM with an RBF kernel with $C = 100$ and $\gamma = 1$. Evaluating using five-fold cross validation, it gave an average accuracy of 94.3% and precision of 96.5%. This compares to a accuracy and precision of 91.6% for a simple non-learning strategy where we assume all pairs of queries from the same IP address within half an hour of each other are in the same query chain. As this difference is relatively small, and computing this feature vector for every query pair is relatively expensive (in particular since it depends on the abstracts retrieved), we decided to rely simply on our heuristic measure. We judged that a precision of over 90% is sufficient for our present purposes. We considered extending the half-hour window on our training data in order to increase the recall, but decided that we were recognizing a sufficient number of query chains without doing so.

However, to gain some insight into the properties of query chains we trained a linear SVM using the same data and computed the total weight on each feature. The features with largest positive weight were CosineDistance(q1, q2), which measures the cosine distance between $q1$ and $q2$, and CosineDistance(doc ids of r1', doc ids of r2'), which measures the overlap between the documents in the top 10 results. This indicates that if two queries are similar, or if they retrieve many of the same documents, then they are more likely to be in the same query chain. The feature with largest negative weight measures the minimum number of results returned by either query normalized between 0 and 1, NormalizedMin($|r1|$, $|r2|$). This indicates that if one of the queries returns few results, the queries are more likely to be in a query chain. Our interpretation is that if $q1$ returns no results, the user is more likely to run a second query.

We conclude that it is possible to segment log files into query chains with reasonable accuracy.

## 7. LEARNING RANKING FUNCTIONS

Given log files recording user behavior on a web search engine, we have shown how to transform the log records into preference judgments in Section 4 after identifying query chains using the method from Section 6. Next, we present an algorithm to learn from these preferences, which we then evaluate using the Osmot search engine described earlier.

We assume as input preference judgments over documents $d_i$ and $d_j$ for a given query $q$ to be of the following form.

$$d_i >_q d_j \qquad (1)$$

Such a preference judgment indicates that $d_i$ is preferred over $d_j$ given $q$. As our retrieval model, we chose a linear retrieval function:

$$rel(d_i, q) = w \cdot \Phi(d_i, q) \qquad (2)$$

where $\Phi(d_i, q)$ (which we define later) is a function that maps documents and queries to a feature vector. Intuitively, it can be thought of as a feature vector describing the quality of the match between a document $d_i$ and the query $q$. $w$ is a weight vector that assigns weights to each of the features in $\Phi$, thus giving us a real valued retrieval function where a higher score indicates a document $d_i$ is estimated to be more relevant to the query $q$. The task of learning a ranking function becomes one of learning an optimal $w$.

### 7.1 Ranking SVMs

We used a modified ranking SVM to learn $w$ in Equation 2. Here, we briefly introduce ranking SVMs [15], which generalize ordinal regression SVMs [13]. We start by rewriting Equation 1 as:

$$w \cdot \Phi(d_i, q) > w \cdot \Phi(d_j, q)$$

We then add a margin, and non-negative slack variables to allow some of the preference constraints to be violated, as is done with classification SVMs. This yields a preference constraint over w.

$$w \cdot \Phi(d_i, q) \geq w \cdot \Phi(d_j, q) + 1 - \xi_{ij}$$

Although we cannot efficiently find a $w$ that minimizes the number of violated constraints, we can minimize an upper bound on the number of violated constraints, $\sum \xi_{ij}$. Simultaneously maximizing the margin leads to the following convex quadratic optimization problem:

$$\begin{aligned} &\min_{w, \xi_{ij}} \tfrac{1}{2} w \cdot w + C \sum_{ij} \xi_{ij} \\ &\quad \text{subject to} \\ &\forall (q, i, j): \ \ w \cdot \Phi(d_i, q) \geq w \cdot \Phi(d_j, q) + 1 - \xi_{ij} \\ &\forall i, j: \ \ \xi_{ij} \geq 0 \end{aligned} \qquad (3)$$

We will later add more constraints to the optimization problem taking advantage of prior knowledge in the learning to rank setting.

## 7.2 Retrieval Function Model

Next we must specify the mapping $\Phi(d_i, q)$. This definition is key in determining what class of ranking functions we can learn, and is therefore particularly important in determining the usefulness of this method. We define two types of features: rank features $\phi^f_{rank}(d, q)$ and term/document features $\phi_{terms}(d, q)$. Rank features serve to exploit the existing retrieval functions $rel^f_0$, while term/document features allow us to learn more fine-grained relationships between particular query terms and specific documents.

First we need a few definitions. Let $T := \{t_1, \ldots, t_N\}$ be all the terms (words) in our dictionary. A query $q$ is a set of terms $q := \{t'_1, \ldots, t'_n\}$ where $t'_i \in T$. Let $D := \{d_1, \ldots, d_M\}$ be the set of all documents in our collection. We assume the original search engine has a number of available retrieval functions $rel^f_0(d, q)$ with $f \in F$. We define $r^f_0(q)$ as the ordered set of results as ranked by $rel^f_0$ for query $q$. In the experiments in this paper, $F$ consists of a single ranking function as provided by Nutch for the sake of simplicity.

Now,

$$\Phi(d, q) = \begin{bmatrix} \phi^{f_1}_{rank}(d, q) \\ \vdots \\ \phi^{f_F}_{rank}(d, q) \\ \phi_{terms}(d, q) \end{bmatrix}$$

$$\phi^f_{rank}(d, q) = \begin{bmatrix} \mathbf{1}(Rank(d \ in \ r^f_0(d, q)) \leq 1) \\ \vdots \\ \mathbf{1}(Rank(d \ in \ r^f_0(q)) \leq 10) \\ \mathbf{1}(Rank(d \ in \ r^f_0(q)) \leq 15) \\ \vdots \\ \mathbf{1}(Rank(d \ in \ r^f_0(q)) \leq 100) \end{bmatrix}$$

$$\phi_{terms}(d, q) = \begin{bmatrix} \mathbf{1}(d = d_1 \wedge t_1 \in q) \\ \vdots \\ \mathbf{1}(d = d_M \wedge t_N \in q) \end{bmatrix}$$

where $\mathbf{1}$ is the indicator function.

Before looking at the term features $\phi_{terms}(d, q)$, let's explore the rank features $\phi^{f_i}_{rank}(d, q)$. For each retrieval function $rel^{f_i}_0$ we have 28 rank features (for ranks 1,2,..,10,15, 20,..,100). Each of these is set to 1 if the rank of the document in $r^{f_i}_0$ is at or above the specified rank.

The rank features allow us to learn weights for the rankings of the original search results. This allows the learned ranking function to combine different retrieval functions with different weights, as is done in prior work described earlier. We do not consider the specific scores assigned by $rel^f_0$ in order to account for potentially different magnitudes of the scores from different retrieval functions. This also ensures that our method could generalize to settings where we do not have access to the scores assigned to documents but only the document ranks. As an example, if some document $d$ is at rank 4 given query $q$ and using retrieval function $f_1$ then $\phi^{f_1}_{rank}(d, q) = [0, 0, 0, 1, \ldots, 1]^T$. If a document is not ranked in the top 100 by the retrieval function $f_1$, then all the features of $\phi^{f_1}_{rank}$ are 0. This means that documents not ranked in the top 100 results by a retrieval function $rel^{f_i}_0$ are indistinguishable using the $\phi^{f_i}_{rank}$ features (although we could increase the maximum rank considered arbitrarily). We chose this cutoff as it is extremely rare for users to look beyond the top 100 results.

We also have $NM$ term/document features. For convenience, let $\phi_{term}^{i,j}(d, q)$ correspond to the term with $d_i$ and $t_j$ in $\phi_{terms}(d, q)$. There is one for every (term, document) pair in $T \times D$. The term/document features allow the ranking function to learn associations between specific query words and documents by assigning a non-zero value to the appropriate weight. This is usually an extremely large number of features, although most never appear in our training data and can thus be ignored. Furthermore, the feature vector $\phi_{terms}(d, q)$ is very sparse. For any particular document $d \in D$, given a query with $|q|$ terms, only $|q|$ of the $\phi_{term}^{i,j}(d, q)$ features are set to 1. Specifically, only the terms for one $i$ value (where $d = d_i$) and with $t_j \in q$ are non-zero. The sparsity makes this problem well suited for solving using support vector machines. A positive value of the weight $w_{term}^{i,j}$, associated with the feature $\phi_{term}^{i,j}$, indicates that $d_i$ is more likely to be relevant to queries containing the term $t_j$, while a negative value means the opposite.

## 7.3  Adding Prior Knowledge

When learning to rank, we have additional prior knowledge that should be incorporated into this problem. Absent any other information, documents with a higher rank in the original ranking should be ranked higher in the learned ranking system. This is intuitive because on average we would expect the document relevance to be a decreasing function of the original rank of the documents, unless the ranking function is particularly poor. We define such additional constraints in this section.

It is also of practical importance to add these constraints: In our training data almost all of the relevance judgments generated state that a lower ranked document is preferred to a higher ranked document. Without additional constraints, a trivial and undesirable solution to the optimization problem in Equation 3 would be one that reverses the original ranking by assigning a negative value to each of the weights corresponding to rank features in $\Phi$. To see this, consider again Figure 4. The "Click $>_{q(q')}$ Skip Above" preferences would be satisfied if the rankings were reversed. These preferences are much more common than "Click First $>_{q(q')}$ No-Click Second" preferences. In the last two preferences classes, the preferred document is also presumably somewhere much lower in the results for $q'$ (if it is not in the results, we can think of it as being at the bottom of the results), and hence the preferences would also be satisfied if the entire ranking were reversed.

We add additional hard constraints to the optimization problem specified in Equation 3. These constraints require that weights for each of the rank features must be greater than a constant positive value $w_{min}$:

$$\forall i \in [1, \ 28|F|]. \ \ w^i \geq w_{min} \qquad (4)$$

Intuitively, $w_{min}$ limits how quickly the original ranking is changed by training data. To see this, briefly consider a setting where we have a single ranking function $f$ and a query $q = t'$ that returns at least 100 results. Let $d_i$ be the document ranked at position $i$ in $r_0^f(q)$. In this case,

$$\phi_{rank}^f(d_{100}, \ q) \ = \ [0, \ \ldots, \ 0, \ 0, \ 1]^T$$
$$\phi_{rank}^f(d_{95}, \ q) \ = \ [0, \ \ldots, \ 0, \ 1, \ 1]^T$$
$$\ldots$$
$$\phi_{rank}^f(d_1, \ q) \ = \ [1, \ \ldots, \ 1, \ 1, \ 1]^T$$

**Figure 6: Two example rankings with four results each, and the combined outputs we would generate by starting with the top ranked document from ranking $r$.**

Calling the part of $w$ that corresponds to rank features $w_{rank}$, from Equation 4 we then get

$$w_{rank} \cdot \phi_{rank}^f(d_{100}, \ q) \ \geq \ w_{min}$$
$$w_{rank} \cdot \phi_{rank}^f(d_{95}, \ q) \ \geq \ 2w_{min}$$
$$\ldots$$
$$w_{rank} \cdot \phi_{rank}^f(d_1, \ q) \ \geq \ 28w_{min}$$

Now say we have a document $d$ that is preferred over $d_1$ but is not in the original results. $d$ would be ranked highest if $rel(d, q) > rel(d_1, q)$. We know from Section 7.2 that only $\phi_{term}^{t', d}(d, q)$ is non-zero in $\phi_{terms}(d, q)$. Expanding and simplifying, this would imply:

$$w_{terms} \cdot \phi_{terms}(d, q) \ \geq \ 28w_{min} + w_{terms} \cdot \phi_{terms}(d_1, q)$$
$$w_{term}^{d, q} \ \geq \ 28w_{min} + w_{terms}^{d_1, q}$$

where $w_{term}^{\alpha, \beta}$ corresponds to $\phi_{term}^{\alpha, \beta}(d, q)$.

The larger $w_{min}$, the larger in magnitude $w_{term}^{d, q}$ and $w_{term}^{d_1, q}$ must be before this happens. A ranking SVM minimizes over $w \cdot w + C \sum \xi_{ij}$, so the terms will only become large if there is sufficient training data to support a reordering.

## 7.4  Evaluation Methodology

In order to evaluate our results, we need an unbiased method for comparing two ranked retrieval functions. For this purpose we use the method detailed in [16]. This method was shown to give an accurate assessment of retrieval quality under reasonable assumptions. Given two ranking functions, we present users with a combination of the results from both. We know that users scan results from top to bottom, so we intertwine the results such that there is no presentation bias favoring either ranking function. This evaluation method is built into the Osmot search engine.

Figure 6 shows two example rankings, $r$ and $r'$, from two different retrieval functions as well as a combination of them, $combined(r, \ r')$. Let $seen(n, r)$ and $seen(n, r')$ be the number of results the user has seen from rankings $r$ and $r'$ respectively after looking at the top $n$ results from the combined ranking. $seen(n, r)$ and $seen(n, r')$ are defined as the smallest number of results that we have to combine from $r$ or $r'$ to produce the top $n$ results of the combined ranking. We generate the combined ranking such that for any $n$, $seen(n, r) \geq seen(n, r') \geq seen(n, r) - 1$. In our example, if the user looks at the top three results in the combined ranking, this is satisfied because $seen(3, r) = 2$ and $seen(3, r') = 2$. If the user looks at the top five results, $seen(5, r) = 4$ and $seen(5, r') = 3$. To compensate for a bias toward the results of $r$ ($seen(n, r)$ is sometimes one bigger than $seen(n, r')$), we randomly switch $r$ and $r'$ half the time. This means that in expectation $seen(n, r) = seen(n, r')$. The property is proved rigorously in [16].

Once we have presented the user with a combined ranking, we need to evaluate which of the two rankings is preferred. We first determine which results the user looked at by taking the lowest ranked clicked-on document as where the user stopped scanning the results (a conservative estimate). If the two rankings are equally good, we would expect the user to click on just as many results from each given that she has seen an equal number from each (in expectation). We measure $clicks(r)$, the number of documents clicked on that are in the top $seen(n, r)$ results of $r$, and similarly $clicks(r')$. For example, in Figure 6, say the user clicked on $d_1$ and $d_5$. We would infer the user looked at the top 3 results. From before, we have $seen(3, r) = seen(3, r') = 2$. Therefore, $clicks(r) = 1$ ($d_1$) and $clicks(r') = 1$ ($d_5$).

If in expectation $clicks(r) > clicks(r')$, we can conclude that the user prefers the ranking $r$ over $r'$. When evaluating ranking functions, we count how often $clicks(r) > clicks(r')$, and $clicks(r) < clicks(r')$. We then use a binomial sign test to verify if the difference in counts of $clicks(r) > clicks(r')$ and $clicks(r) < clicks(r')$ is statistically significant. If so, we can say one ranking is preferred over the other.

## 7.5 Training the Ranking SVM

We collected training data from the CUL search engine using the original ranking function between June and December 2004. During this time, we recorded user queries and clicks, observing 9,949 queries and 7,429 clicks. While we were collecting this data, the users saw results as ranked by the built-in Nutch retrieval function, which we denote as $rel_0$. This gave 120,134 preferences constraints by applying all six strategies introduced above. We call these preferences $P_{QC}$. Of these, 45,610 preferences were generated without using the query chain strategies. We call this subset of the preferences $P_{NC}$.

After adding the hard constraints as described above, we trained a ranking SVM for each of the two sets of preferences with a linear kernel and a default value of C using $SVM^{light}$[14]. We set $w_{min} = 1$. Using the preferences $P_{QC}$ we learned a retrieval function $rel_{QC}$ and using the preferences $P_{NC}$ we learned $rel_{NC}$. The former model has 41,354 support vectors, while the latter has 18,034.

The ranking model learned using query chains, $rel_{QC}$, instantiated 18,748 features. The number of features instantiated can be expected to grow almost linearly in the size of the document collection, and sub-linearly in the amount of training data collected (depending on overall user search behavior). However, this did not pose a problem from the SVM solver because all the preference judgments were sparse.

| Evaluation | User Prefers | | |
|---|---|---|---|
| Mode | Chains | Other | Indifferent |
| $rel_{QC}$ vs. $rel_0$ | 392 (32%) | 239 (20%) | 579 (47%) |
| $rel_{QC}$ vs. $rel_{NC}$ | 211 (17%) | 160 (13%) | 855 (70%) |

**Table 3: Results on Cornell Library search engine. $rel_0$ is the original retrieval function, $rel_{QC}$ is that trained using query chains, and $rel_{NC}$ is that trained without using query chains.**

## 7.6 Results and Discussion

We evaluated the ranking functions on the CUL search from 10 December 2004 through 18 February 2005 using the evaluation method described in Section 7.4. When a user connected to the search engine, we randomly selected an "evaluation mode" for that user. The user either saw a ranking combining $rel_0$ and $rel_{QC}$ or a ranking combining $rel_{QC}$ and $rel_{NC}$. For consistency, we kept the same combination for the duration of each user's session (otherwise, if the user immediately re-ran the same query he or she may confusingly get different results).

During the evaluation, we collected about 1200 queries in each evaluation mode. The results for both evaluation modes are shown in Table 3. These results show a number of interesting properties. Firstly, 53% of the time $rel_{QC}$, the ranking function trained using query chains, performs differently to the original ranking function, $rel_0$. 30% of the time the two trained ranking functions perform differently. In particular, the first of these values indicates that our method often makes a difference in search engine performance. Given that the original ranking function is reasonable, it would be surprising if these values were much higher. As long as our method does not cause relevant documents that are ranked highly by $rel_0$ to be lowered in rank, we would see identical performance in the cases when $rel_0$ performs well.

Secondly, from Table 3 we see that $rel_{QC}$ outperforms $rel_0$ more often than we would expect at random were the two ranking functions equally good. Using a binomial sign test, and the null hypothesis that the two ranking functions are equally effective, we are able to reject the null hypothesis with over 99% confidence. This establishes that our learned ranking function is an improvement over the original one. Of course, given the new ranking function, we are collecting new training data and can re-run the whole learning process. We expect this to produce continued improvement in ranking performance.

Finally, the model trained using query chains outperforms the model trained without using query chains with over 99% confidence, using the same test. This demonstrates that by exploiting the information about query chains present in log files, we are able to see a measurable additional improvement in search engine performance over what we would see without using this extra information.

One may wonder if it makes sense to learn associations between specific query words and documents. Given our initial 9,949 training queries, Table 4 shows the top ten words that appear most frequently in queries. We see that queries tend to be repetitive. Ignoring the three stopwords in the top ten words, we found that at least one of the remaining seven words appears in 12% of all queries. At least one of the top 100 words (removing stopwords) appears in 38% of all queries. Moreover, for many popular queries, there appear to be only a few documents that are truly relevant to the query. Hence it is not surprising that by learning individual query word/document associations we can see significant improvements in ranking results.

| Word | Fraction of queries |
|------|---------------------|
| of | 3.56 % |
| library | 2.75 % |
| bibliography | 2.60 % |
| and | 2.55 % |
| annotated | 2.42 % |
| reserve | 2.32 % |
| citation | 1.99 % |
| web | 1.48 % |
| the | 1.41 % |
| course | 1.33 % |

**Table 4: The most common words to appear in queries in the training data, and the fraction of queries in which they occur.**

| Word | Document | Weight |
|------|----------|--------|
| lexus | Lexis-Nexis Academic Universe | 22.8 |
| ebook | CUL eContent Collection | 22.5 |
| reuleaux | CUL Digital Collections | 21.8 |
| and | Printable News and Notes 07/03 | 19.6 |
| oed | Dictionaries and Encyclopedias | 19.5 |
| ndlf | Management meeting notes 03/97 | -21.0 |
| ndlf | Management meeting notes 02/97 | -20.6 |
| ndlf | Management meeting notes 04/96 | -19.5 |
| ndlf | Management meeting notes 04/96 | -18.6 |
| instruction | Library Research Workshops | -18.3 |

**Table 5: Five most positive and most negative feature weights in the ranking function learned using query chains on the Cornell University Library (CUL) search engine**

In order to understand where the improvements are coming from, it is useful to look at the word/document features with largest positive and negative weights. The top and bottom five features are given in Table 5. First we consider the top five features, which for the most part describe very sensible associations. The feature for "lexus" is associated with the main homepage of the Lexis-Nexis library resource. This is clearly a spelling correction, with a search for "lexus" originally returning no results. The same search now places the correct document at the top of the results. The feature for "ebook" returns the main ebooks web page. A search for ebook previously returned seven results, none of which were particularly useful. The top one, titled "Answers to Frequent Job Searching Research Questions", happened to mention access to ebooks from off campus. The feature for "reuleaux" is associated with an FAQ page page about the CUL digital collections. The web page provides a clear link to a site that describes models designed by Professor Reuleaux. This contrasts with the original top result being a broken link, and the second result being a newsletter with only passing reference to the model collection. The feature for "and" is of little practical interest (we did not remove stopwords). Finally, the fifth word "oed" is an acronym for the "Oxford English Dictionary". The associated document clearly links to it, in contrast with the original top result which was an information bulletin showing a set of screen shots how to get to the OED among other things.

The five features with the most negative weights in Table 5 are equally interesting. Four of them relate to meeting notes mentioning the National Digital Library Foundation. Using the original ranking function, this search generated just 6 results with only such meeting notes. With the learned system, a search for "ndlf" now returns similar results to a search for "National Digital Library Foundation". These results appear slightly more useful from the short abstracts that are presented. However, we discovered that in fact the search engine had not indexed the main NDLF web page. We see here that the search system has recognized users running chains of queries looking for the NDLF website, although none have been successful in finding it. Despite this, some of the worst results for this query have indeed been pushed down the results list. The fifth feature is harder to interpret, but from log files it appears that users looking for the Department of Learning and Instruction saw this result and repeatedly skipped it. This document used to appear as the top result given the query "instruction".

# 8. CONCLUSIONS AND FUTURE WORK

In this paper, we have demonstrated that query chains can be used to extract useful information from search engine log files. After presenting an algorithm to infer preference judgments from log files, we showed that the preferences judgments are valid, independent of the learning method. We then presented a method to identify query chains, and an algorithm that uses the preference judgments to learn an improved ranking function. The model used for the ranking function is more general than in previous work. In particular, it allows the algorithm to learn to include new documents originally not present in initial search results in the learned rankings. The evaluation shows our approach to be effective, and that it can learn highly flexible modifications to the original search results. The Osmot search engine is available to the research community[4].

A natural question that arises in this setting is the tolerance of this method to noise in the training data, particularly should users click in malicious ways. While we used noisy real-world data, we plan to explicitly study the effect of noise, words with two meanings, and click-spam on our approach.

Also, the strategies presented in Section 4.1 give equal weight to each pair of queries within a query chain. However, we suspect that there is additional information present in the position of a query within a chain, and of a click within the sequence of all clicks for each chain. In particular, it is possible that the last query and last clicks may be more informative than earlier ones.

Thirdly, exploiting the fact that it is possible to collect virtually unlimited amounts of search engine log data, we believe that the methods presented in this paper can be extended to learn personalized ranking functions. We are currently refining the Osmot search engine and will use it on the arXiv.org e-Print archive[5] in order to conduct such experiments.

Finally, from a practical perspective our approach pushes the limit of problems that current SVM implementations can solve in reasonable time due to the number of constraints we generate. We believe there is room for improvement in learning methods to efficiently deal with such large numbers of constraints, for example using an incremental optimization approach. Perhaps there are also alternative learning methods, rather than SVMs, that can be used to optimize over preference constraints while being able to learn sufficiently general ranking functions.

---

[4]http://www.cs.cornell.edu/~filip/osmot/
[5]http://www.arxiv.org/

# 9. ACKNOWLEDGMENTS

We would first like to thank three undergraduate and Masters students, Charitha Tillekeratne, Alex Cheng and Atit Patel for working on an initial implementation of the CUL search engine. We also thank our colleagues Laura Granka, Bing Pang, Helene Hembrooke and Geri Gay for their collaboration in the eye tracking study. We thank Paul Houle for his help and support regarding the full text search engine for the Cornell University Library web pages, and Simeon Warner and Paul Ginsparg for interesting discussions on this topic. We also thank the subjects of the eye tracking study and the five relevance judges. This work was funded under NSF CAREER Award IIS-0237381 and the KD-D grant.

# 10. REFERENCES

[1] B. Bartell and G. W. Cottrell. Learning to retrieve information. In *Proceedings of the Swedish Conference on Connectionism*, 1995.

[2] B. T. Bartell, G. W. Cottrell, and R. K. Belew. Automatic combination of multiple ranked retrieval systems. In *Annual ACM Conference on Research and Development in Information Retrieval (SIGIR)*, 1994.

[3] D. Beeferman and A. Berger. Agglomerative clustering of a search engine query log. In *ACM International Conference on Knowledge Discovery and Data Mining (KDD)*, 2000.

[4] J. Boyan, D. Freitag, and T. Joachims. A machine learning architecture for optimizing web search engines. In *AAAI Workshop on Internet Based Information Systems*, August 1996.

[5] A. Broder. A taxonomy of web search. *SIGIR Forum*, 26(2):3–10, 2002.

[6] W. W. Cohen, R. E. Shapire, and Y. Singer. Learning to order things. *Journal of Artificial Intelligence Research*, 10:243–270, 1999.

[7] K. Crammer and Y. Singer. Pranking with ranking. In *Proceedings of the Conference on Neural Information Processing Systems (NIPS)*, 2001.

[8] S. Cucerzan and E. Brill. Spelling correction as an iterative process that exploits the collective knowledge of web users. In *Proceedings of Conference on Empirical Methods in Natural Language Processing (EMNLP)*, pages 293–300, 2004.

[9] Y. Freund, R. Iyer, R. E. Schapire, and Y. Singer. An efficient boosting algorithm for combining preferences. In *International Conference on Machine Learning (ICML)*, 1998.

[10] G. W. Furnas. Experience with an adaptive indexing scheme. In *Proceedings of Conference on Human Factors in Computing Systems (CHI)*, pages 131–135, 1985.

[11] L. Granka. Eye tracking analysis of user behaviors in online search. Master's thesis, Cornell University, 2004.

[12] L. Granka, T. Joachims, and G. Gay. Eye-tracking analysis of user behavior in www search. In *Poster Abstract, Proceedings of the Conference on Research and Development in Information Retrieval (SIGIR)*, 2004.

[13] R. Herbrich, T. Graepel, and K. Obermayer. Large margin rank boundaries for ordinal regression. In A. S. et al., editor, *Advances in Large Margin Classifiers*, pages 115–132, 2000.

[14] T. Joachims. Making large-scale SVM learning practical. In B. Schlkopf, C. Burges, and A. Smola, editors, *Advances in Kernel Methods – Support Vector Machines*. MIT Press, 1999.

[15] T. Joachims. Optimizing search engines using clickthrough data. In *Proceedings of the ACM Conference on Knowledge Discovery and Data Mining (KDD)*. ACM, 2002.

[16] T. Joachims. Evaluating retrieval performance using clickthrough data. In J. Franke, G. Nakhaeizadeh, and I. Renz, editors, *Text Mining*, pages 79–96. Physica/Springer Verlag, 2003.

[17] T. Joachims, L. Granka, B. Pang, H. Hembrooke, and G. Gay. Accurately interpreting clickthrough data as implicit feedback. In *Annual ACM Conference on Research and Development in Information Retrieval (SIGIR)*, 2005.

[18] R. Jones and D. C. Fain. Query word deletion prediction. In *Annual ACM Conference on Research and Development in Information Retrieval (SIGIR)*, pages 435–436, 2003.

[19] C. Kemp and K. Ramamohanarao. Long-term learning for web search engines. In T. E. et al., editor, *PKDD*, pages 263–274, 2003.

[20] T. Lau and E. Horvitz. Patterns of search: Analyzing and modelling web query refinement. In *Proceedings of the 7th International Conference on User Modeling*, 1999.

[21] B. U. Oztekin, G. Karypis, and V. Kumar. Expert agreement and content based reranking in a meta search environment using Mearf. In *Proceedings of the 11th International Conference on the World Wide Web (WWW)*, pages 333–344, 2002.

[22] S. Rajaram, A. Garg, Z. S. Zhou, and T. S. Huang. Classification approach towards ranking and sorting problems. In *Lecture Notes in Artificial Intelligence*, volume 2837, pages 301–312, September 2003.

[23] F. Scholer and H. E. Williams. Query association for effective retrieval. In *Proceedings of the 11th International Conference on Information and Knowledge Management*, pages 324–331, 2002.

[24] C. Silverstein, M. Henzinger, H. Marais, and M. Moricz. Analysis of a very large AltaVista query log. Technical Report 1998-014, Digital SRC, 1998.

[25] Q. Tan, X. Chai, W. Ng, and D.-L. Lee. Applying co-training to clickthrough data for search engine adaptation. In *Proceedings of the 9th International Conference on Database Systems for Advanced Applications (DASFAA)*, 2004.